# Kinematic Analysis of a Family of 3R Manipulators

Maher Baili, Philippe Wenger and Damien Chablat
Institut de Recherche en Communications et Cybernétique de Nantes, UMR C.N.R.S. 6597
1, rue de la Noë, BP 92101, 44321 Nantes Cedex 03 France

*Abstract*— *The workspace topologies of a family of 3-revolute (3R) positioning manipulators are enumerated. The workspace is characterized in a half-cross section by the singular curves. The workspace topology is defined by the number of cusps that appear on these singular curves. The design parameters space is shown to be divided into five domains where all manipulators have the same number of cusps. Each separating surface is given as an explicit expression in the DH-parameters. As an application of this work, we provide a necessary and sufficient condition for a 3R orthogonal manipulator to be cuspidal, i.e. to change posture without meeting a singularity. This condition is set as an explicit expression in the DH parameters.*

I. INTRODUCTION

A positioning manipulator may be used as such for positioning tasks in the Cartesian space or as the regional structure of a 6R manipulator with spherical wrist. Most industrial regional structures have the same kinematic architecture, namely, a vertical revolute joint followed by two parallel joints, like the Puma. Such manipulators are always *noncuspidal* (i.e. must meet a singularity to change their posture) and they have four inverse kinematic solutions (IKS) for all points in their workspace (assuming unlimited joints). This paper focuses on alternative manipulator designs, namely, positioning 3R manipulators with orthogonal joint axes (orthogonal manipulators). Orthogonal manipulators may have different global kinematic properties according to their link lengths and joint offsets. They may be *cuspidal*, that is, they can change their posture without meeting a singularity [1, 2]. Cuspidal robots were unknown before 1988 [3], when a list of conditions for a manipulator to be noncuspidal was provided [4, 5]. This list includes simplifying geometric conditions like parallel and intersecting joint axes [4] but also nonintuitive conditions [5]. A general necessary and sufficient condition for a 3-DOF manipulator to be cuspidal was established in [6], namely, the existence of at least one point in the workspace where the inverse kinematics admits three equal solutions. The word "cuspidal manipulator" was defined in accordance to this condition because a point with three equal IKS forms a cusp in a cross section of the workspace [4, 7]. The categorization of all generic quaternary 3R manipulators was established in [8] based on the homotopy class of the singular curves in the joint space. [9] proposed a procedure to take into account the cuspidality property in the design process of new manipulators. More recently, [10] applied efficient algebraic tools to the classification of 3R orthogonal manipulators with no offset on their last joint. Five surfaces were found to divide the parameters space into 105 cells where the manipulators have the same number of cusps in their workspace. The equations of these five surfaces were derived as polynomials in the DH-parameters using Groebner Bases. A kinematic interpretation of this theoretical work was conducted in [11] : the authors analyzed general kinematic properties of one representative manipulator in each cell. Only five different cases were found to exist. However, the classification in [11] did not provide the equations of the separating surfaces in the parameters space for the five cells associated with the five cases found. The purpose of this work is to classify a family of 3R positioning manipulators according to the topology of their workspace, which is defined by the number of cusps appear on the singular curves. The design parameters space is shown to be divided into five domains where all manipulators have the same number of cusps. As an application of this work, a necessary and sufficient condition for a 3R orthogonal manipulator to be cuspidal is provided as an explicit expression in the DH parameters. This study is of interest for the design of new manipulators.

The rest of this article is organized as follows. Next section presents the manipulators under study and recalls some preliminary results. The classification is established in section III. Section IV states the necessary and sufficient condition and section V concludes this paper.

II. MANIPULATOR UNDER STUDY

The manipulators studied in this paper are orthogonal with their last joint offset equal to zero. The remaining lengths parameters are referred to as $d_2$, $d_3$, $d_4$, and $r_2$ while the angle parameters $\alpha_2$ and $\alpha_3$ are set to –90° and 90°, respectively. The three joint variables are referred to as $\theta_1$, $\theta_2$ and $\theta_3$, respectively. They will be assumed unlimited in this study. Figure 1 shows the kinematic architecture of the manipulators under study in the zero configuration. The position of the end-tip (or wrist center) is defined by the Cartesian coordinates $x$, $y$ and $z$ of the operation point $P$ with respect to a reference frame (O, **x**, **y**, **z**) attached to the manipulator base as shown in Fig. 1.

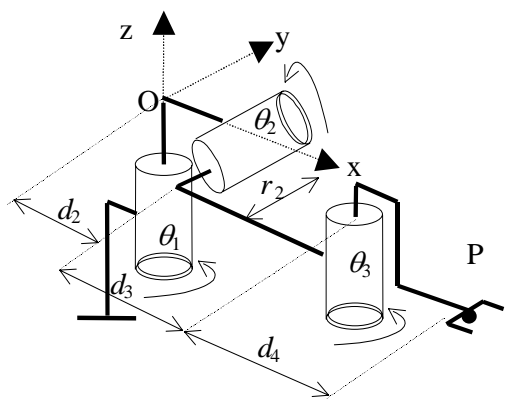

Figure 1 : Orthogonal manipulators under study.

The singularities of general 3R manipulators can be determined by calculating the determinant of the Jacobian matrix. For the orthogonal manipulators under study, the determinant of the Jacobian matrix takes the following form [15]:

$$\det(\mathbf{J}) = (d_3 + c_3 d_4)(s_3 d_2 + c_2(s_3 d_3 - c_3 r_2)) \tag{1}$$

where $c_i=\cos(\theta_i)$ and $s_i=\sin(\theta_i)$. A singularity occurs when $\det(\mathbf{J})=0$. Since the singularities are independent of $\theta_1$, the contour plot of $\det(\mathbf{J})=0$ can be displayed as curves in $-\pi \leq \theta_2 < \pi, -\pi \leq \theta_3 < \pi$. The singularities can also be displayed in the Cartesian space by plotting the points where the inverse kinematics has double roots [13]. Thanks to their symmetry about the first joint axis, it is sufficient to draw a half cross-section of the workspace by plotting the points ($\rho = \sqrt{x^2 + y^2}$, $z$).

If $d_3 > d_4$, the first factor of $\det(\mathbf{J})$ cannot vanish and the singularities form two distinct curves $S_1$ and $S_2$ in the joint space [15]. When the manipulator is in such a singularity, there is line that passes through the operation point and that cuts all joint axes [4]. The singularities form two disjoint sets of curves in the workspace. These two sets define the internal boundary $WS_1$ and the external boundary $WS_2$, respectively, with $WS_1=f(S_1)$ and $WS_2=f(S_2)$. Figure 2(a) shows the singularity curves when $d_3=2$, $d_4=1.5$ and $r_2=1$. For this manipulator, the internal boundary $WS_1$ has four cusp points, where three IKS coincide. It divides the workspace into one region with two IKS (the outer region) and one region with four IKS (the inner region).

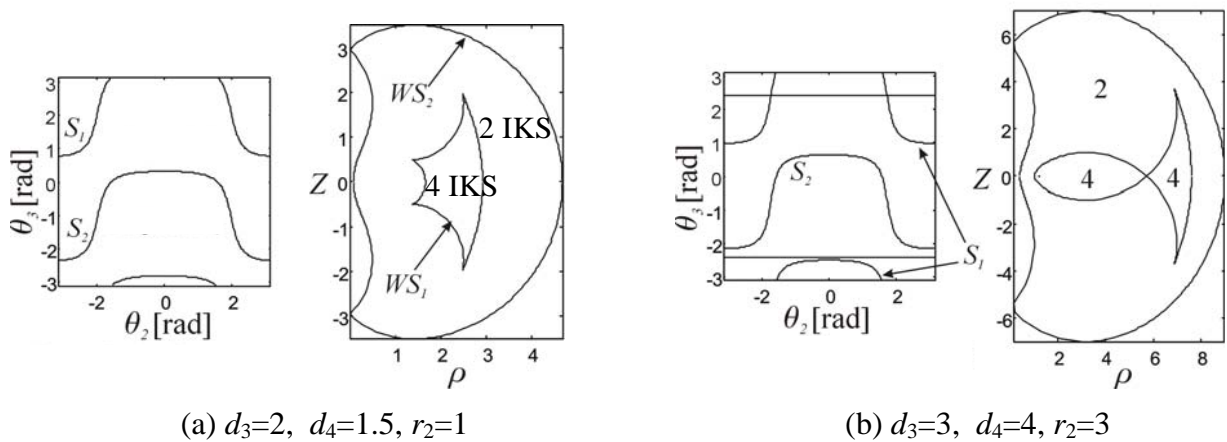

(a) $d_3=2$, $d_4=1.5$, $r_2=1$          (b) $d_3=3$, $d_4=4$, $r_2=3$

Figure 2 : Singularity curves when $d_3>d_4$ (a) and when $d_3<d_4$ (b)

If $d_3 \leq d_4$, the operation point can meet the second joint axis whenever $\theta_3 = \pm\arccos(-d_3/d_4)$ and two horizontal lines appear in the joint space. No additional curve appears in the workspace cross-section but only two points. This is because, since the operation point meets the second joint axis when $\theta_3 = \pm\arccos(-d_3/d_4)$, the location of the operation point does not change when $\theta_2$ is rotated. Figure 2 (*b*) shows the singularity curves of a manipulator such that $d_3=3$, $d_4=4$, $r_2=3$.

### III. WORKSPACE TOPOLOGIES

The workspace is defined by the topology of the singular curves, which we characterize by the number of cusps. A cusp is associated with one point with three equal IKS. These singular points are interesting features for characterizing the workspace shape and the accessibility in the workspace.

For now on and without loss of generality, $d_2$ is set to 1. Thus, we need handle only three parameters $d_3$, $d_4$ and $r_2$. Efficient computational algebraic tools were used in [10] to provide the equations of five separating surfaces, which were shown to divide the parameter space into 105 cells. But [11] showed that only 5 cells should exist, which means that one or more surfaces among the five ones found in [10] are not relevant. However, [11] did not try to find which surfaces are really separating. To derive the equations of the true separating surfaces, we need to investigate the transitions between the five cases. First, let us recall the five different cases found in [11]. The first case is a binary manipulator (i.e. it has only two IKS) with no cusp and a hole (Fig. 3). The remaining four cases are quaternary manipulators (i.e. with four IKS). The second case is a manipulator with four cusps on the internal boundary. Figure 4 shows a manipulator of this case with a hole. Transition between case 1 and case 2 is a manipulator with two points with four equal IKS, where one node and two cusps coincide [15].

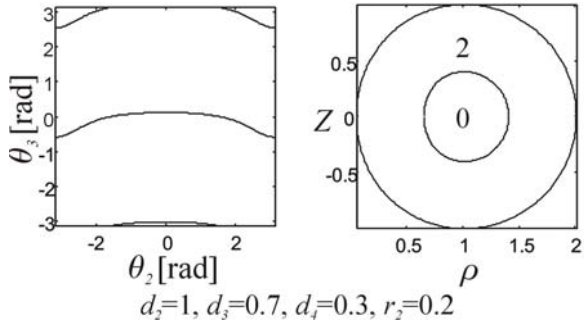
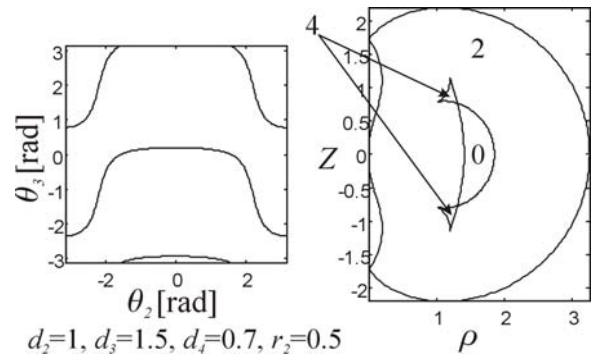

Figure 3 : Manipulator of case 1        Figure 4 : Manipulator of case 2

Deriving the condition for the inverse kinematic polynomial to have four equal roots yields the equation of the separating surface [15]

$$d_4 = \sqrt{\frac{1}{2}\left(d_3^2 + r_2^2 - \frac{(d_3^2 + r_2^2)^2 - d_3^2 + r_2^2}{AB}\right)} \quad (2)$$

where

$$A = \sqrt{(d_3+1)^2 + r_2^2} \text{ and } B = \sqrt{(d_3-1)^2 + r_2^2} \qquad (3)$$

Note that there exist two other instances of case 2: the manipulator shown in Fig. 2a with no hole, and a manipulator where the upper and lower segments of the internal boundary cross, forming a '2-tail fish' [15].

The third case is a manipulator with only two cusps on the internal boundary, which looks like a fish with one tail (Fig. 2b). As shown in [15], transition between case 2 and case 3 is characterized by a manipulator for which the singular line given by $\theta_3 = -\arccos(-d_3/d_4)$ is tangent to the singularity curve $S_1$. Expressing this condition yields the equation of the separating surface, where $A$ is given by (3):

$$d_4 = \frac{d_3}{1+d_3} \cdot A \qquad (4)$$

The fourth case is a manipulator with four cusps. Unlike case 2, the cusps are not located on the same boundary (Fig. 5). Transition between case 3 and case 4 is characterized by a manipulator for which the singular line given by $\theta_3 = -\arccos(-d_3/d_4)$ is tangent to the singularity curve $S_2$ [15]. Expressing this condition yields the equation of the separating surface, where $B$ is given by (3):

$$d_4 = \frac{d_3}{d_3 - 1} \cdot B \text{ and } d_3 > 1 \qquad (5)$$

Last case is a manipulator with no cusp. Unlike case 1, the internal boundary does not bound a hole but a region with 4 IKS. The two isolated singular points inside the inner region are associated with the two singularity lines. Transition between case 4 and case 5 is characterized by a manipulator for which the singular line given by $\theta_3 = +\arccos(-d_3/d_4)$ is tangent to the singularity curve $S_1$ [15]. Expressing this condition yields the equation of the separating surface:

$$d_4 = \frac{d_3}{1-d_3} \cdot B \text{ and } d_3 < 1 \qquad (6)$$

where $B$ is given by (3).

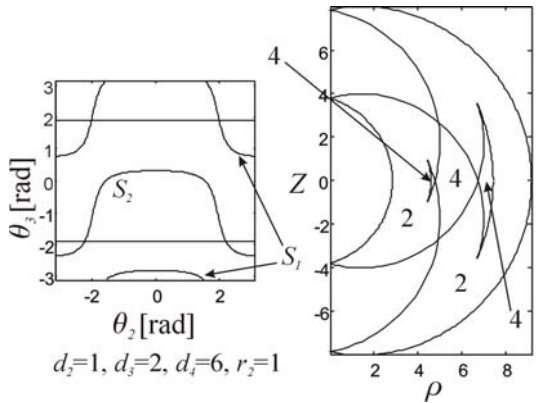
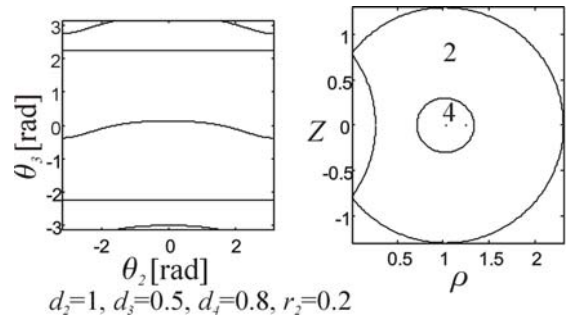

Figure 5: Manipulator of case 4     Figure 6 : Manipulator of case 5

We have provided the equations of four surfaces that divide the parameters space into five domains where the number of cusps is constant. Figure 7 shows the plots of these surfaces in a section ($d_3$, $d_4$) of the parameter space for $r_2=1$. Domains 1, 2, 3, 4 and 5 are associated with manipulators of case 1, 2, 3, 4 and 5, respectively. $C_1$, $C_2$, $C_3$ and $C_4$ are the right hand side of (2), (4), (5) and (6), respectively.

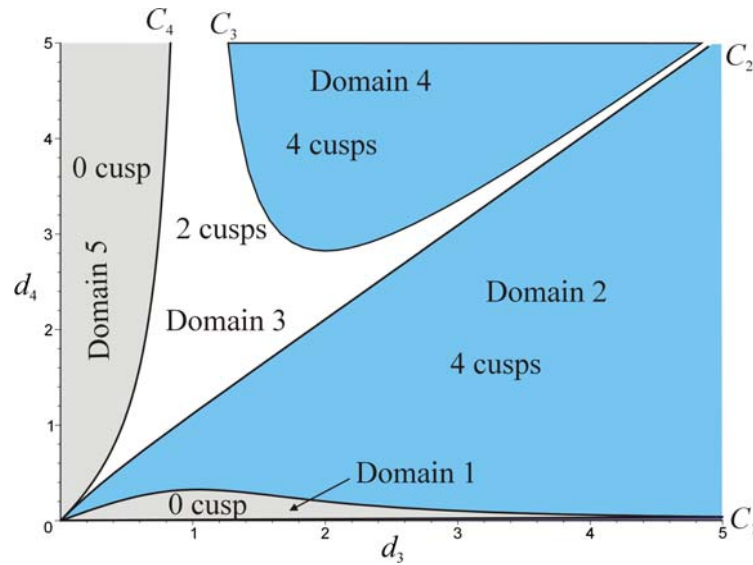

Fig. 7 : Plots of the four separating surfaces in a section ($d_3$, $d_4$) of the parameter space for $r_2=1$.

IV. NECESSARY AND SUFFICIENT CONDITION FOR A MANIPULATOR TO BE CUSPIDAL

The above classification provides a means to derive an explicit DH parameter condition for an orthogonal manipulator to be cuspidal, i.e., to change posture without meeting a singularity. In effect, as shown in Fig. 7, any cuspidal manipulator belongs to domains 2, 3 or 4. Thus, the DH-parameters must satisfy $d_4 > C_1$ and ($d_4 < C_4$ or $d_3 > 1$). Thus, a necessary and sufficient condition for an orthogonal manipulator to be cuspidal is (by dividing the parameters by $d_2$, one gets the general formula for manipulators such that $d_2 \neq 1$)

and
$$d_4 > \sqrt{\frac{1}{2}\left(d_3^2 + r_2^2 - \frac{(d_3^2 + r_2^2)^2 - (d_3^2 + r_2^2)d_2^2}{\sqrt{(d_3 + d_2)^2 + r_2^2}\sqrt{(d_3 - d_2)^2 + r_2^2}}\right)}$$
$$d_3 \geq d_2 \text{ or } \left(d_3 < d_2 \text{ and } d_4 < \frac{d_3}{d_2 - d_3}\sqrt{(d_3 - d_2)^2 + r_2^2}\right) \quad (7)$$

This condition is explicit and can be checked very easily.

V. CONCLUSION

A family of 3R manipulators was classified according to the topology of the workspace, which was defined as the number of cusps. The design parameters space was shown to be divided into five domains where all manipulators have the same number of cusps. Each separating surface was given as an explicit expression in the DH-parameters. An interesting application result of this work is the establishment of a necessary and sufficient condition for a manipulator to be cuspidal, i.e., to change posture without meeting a singularity. This condition was set as an explicit expression in the DH parameters.

Kinematic analysis of a family of 3R manipulators, IFToMM, Problems of Mechanics, M. Baili, P. Wenger and D. Chablat, Vol. 15, N°2, pp 27-32, juillet 2004.

# SUMMARY

## Kinematic Analysis of a Family of 3R Manipulators

Maher Baili, Philippe Wenger and Damien Chablat
Institut de Recherche en Communications et Cybernétique de Nantes, u.m.r. C.N.R.S. 6597
1, rue de la Noë, BP 92101, 44321 Nantes Cedex 03 France

*Abstract—* The workspace topologies of a family of 3-revolute (3R) positioning manipulators are enumerated. The workspace is characterized in a half-cross section by the singular curves. The workspace topology is defined by the number of cusps that appear on these singular curves. The design parameters space is shown to be divided into five domains where all manipulators have the same number of cusps. Each separating surface is given as an explicit expression in the DH-parameters. As an application of this work, we provide a necessary and sufficient condition for a 3R orthogonal manipulator to be cuspidal, i.e. to change posture without meeting a singularity. This condition is set as an explicit expression in the DH parameters.

*Keywords—Workspace, Singularity, 3R manipulator, Cuspidal manipulator.*